%% file: main.tex
\documentclass{article}

\usepackage{times}
\usepackage{graphicx} %
\usepackage{subfigure} 

\usepackage{natbib}

\usepackage{algorithm}
\usepackage{algorithmicx}
\usepackage{algpseudocode}
\newcommand{\algcomment}[1]{\Comment{\textit{#1}}}
\algnewcommand{\LineComment}[1]{\(\triangleright\) \textit{#1}}

\usepackage{microtype}      %
\usepackage{amsmath}
\usepackage{graphicx}
\usepackage{color}
\usepackage{booktabs}       %
\usepackage{amsfonts}       %
\usepackage{nicefrac}       %

\input{commands}

\newcommand{\child}{\mathsf{child}}
\newcommand{\treepath}{\mathsf{path}}
\newcommand{\parent}{\mathsf{parent}}
\newcommand{\sieb}{\mathsf{sibling}}
\newcommand{\leaf}{\mathsf{final}}
\newcommand{\ftheta}{f_{\theta}}
\newcommand{\knn}{$k$-NN}
\usepackage{hyperref}

\usepackage[nohyperref, accepted]{icml2017}

\icmltitlerunning{Differentiable Boundary Trees}

\begin{document} 

\twocolumn[
\icmltitle{Learning Deep Nearest Neighbor Representations Using \\ Differentiable Boundary Trees}

\icmlsetsymbol{equal}{*}

\begin{icmlauthorlist}
\icmlauthor{Daniel Zoran}{to}
\icmlauthor{Balaji Lakshminarayanan}{to}
\icmlauthor{Charles Blundell}{to}
\end{icmlauthorlist}

\icmlaffiliation{to}{DeepMind, London,  UK}

\icmlcorrespondingauthor{Daniel Zoran}{danielzoran@google.com}

\icmlkeywords{Nearest Neighbor, Boundary Tree}%

\vskip 0.3in
]

\printAffiliationsAndNotice{}  %

\begin{abstract}
Nearest neighbor (\knn) methods have been gaining popularity in recent years in light of advances in hardware and efficiency of algorithms. There is a plethora of methods to choose from today, each with their own advantages and disadvantages. One requirement shared between all \knn \ based methods is the need for a good representation and distance measure between samples.

We introduce a new method called \emph{differentiable boundary tree} which allows for learning deep \knn\ representations. We build on the recently proposed \emph{boundary tree} algorithm which allows for efficient nearest neighbor classification, regression and retrieval. By modelling traversals in the tree as stochastic events, we are able to form a \emph{differentiable} cost function which is associated with the tree's predictions. Using a deep neural network to transform the data and back-propagating through the tree allows us to learn good representations for \knn\ methods. 

We demonstrate that our method is able to learn suitable representations allowing for very efficient trees with a clearly interpretable structure.

\end{abstract}

\section{Introduction}
There has been a growing interest in $k$-nearest neighbor (\knn) based methods in recent years \citep{muja2009fast,weinberger2005distance}. With the increase in computational power,
\knn\ based methods have become viable for many different problems and have been used in various different contexts such as classification,
regression and retrieval \citep{friedman1977algorithm, beygelzimer2006cover}.

One challenge that all nearest neighbor methods share is finding good representations and distance measures between samples. This can make all the difference to the success of a given \knn\ method, and more often
than not, the representation is chosen ad-hoc or engineered by hand.

Another challenge that \knn\ based methods pose is their computational and memory requirements. As much as hardware has advanced, this still remains a big problem. Typically \knn\ methods scale quite badly with data size, often with challenging trade-offs -- faster retrieval usually requires
more memory \citep{friedman1977algorithm}.

The \emph{Boundary Tree (or forest)} algorithm, recently proposed by \citep{BF}, has some very appealing properties in tackling the computational and memory requirements of \knn\  methods. In a nutshell, a boundary tree is a tree where each node corresponds to a training instance. At query time, the current node is to be the root node and a test data point is compared to the current node and its children: if the current node is the closest data point, the prediction is the class
label or regression target at that node. Otherwise, the process recurses along the closest child. Hence, a boundary tree might be thought of as a hierarchical data structure that enables efficient \knn\  queries. A boundary forest is an ensemble of randomized boundary trees. Boundary trees allow for fast \knn\  classification, regression and retrieval and their memory requirements grow very slowly with the amount of data presented, all within a simple and elegant formulation.

However, like all \knn\  based methods, boundary trees still require a good representation and distance measure in order to perform well. \citet{BF} used $\Ltwo$ distance with raw features between inputs as their distance measure. While the $\Ltwo$ distance might be a sensible metric on pre-processed features (e.g.~SIFT and HOG), it is clearly inadequate when dealing with raw (pixel) inputs.

In this paper we show how to learn useful representations for \knn\  methods in an \emph{end-to-end} fashion, by deriving a differentiable cost function for boundary trees. 
Given such a differentiable function we are able to use the power of deep neural networks to learn good representation of the data by back-propagating into the tree and stored samples. We demonstrate the effectiveness of the methods on the MNIST data set and provide further analysis into what boundary trees are capable of given a good learned representation. We discuss the capabilities and limitations of the method and show show that with a good representation, boundary trees keep either ``prototypical" examples of the data in addition to, as their name suggest, boundary cases -- resulting in an interpretable and efficient data structure which still allows good classification results.

\section{Model}
\subsection{Boundary Trees}

\begin{figure*}[ht]
\centering
\includegraphics[width=0.99\linewidth]{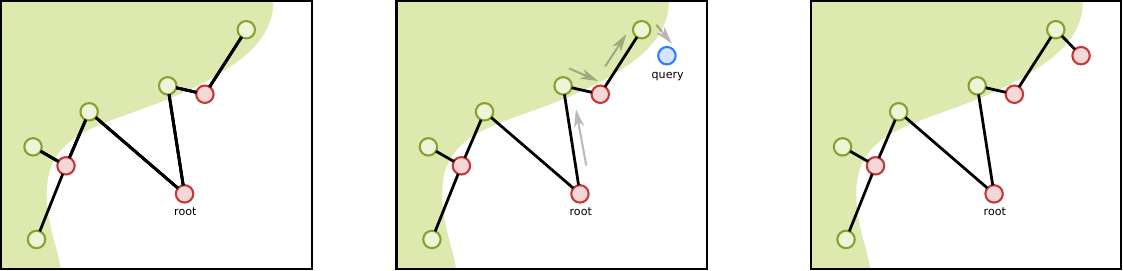}
\caption{\small Boundary trees are built in an online manner, sample by sample. From left to right: Given the current tree (depicted in the left image) we start with the root node. For each query we recursively traverse the tree, choosing the locally closest node to the query node at each step. Once traversal stops we use the final node's class to make the prediction (middle image). If the prediction is wrong (as it is in this example) we add the query node as a child to the final node, resulting in a new tree (depicted in the right image). Otherwise the query node is discarded. Edges in the tree cross class boundaries by definition and samples will tend to reside close to the boundaries, hence the name "Boundary Tree".}
\label{fig:boundary_trees_intro}
\end{figure*}

Since boundary trees are new and relatively unknown, we  provide a brief introduction to this elegant method here. We refer the reader to \citep{BF} for a complete description of the algorithm. We  only deal with the case of classification although regression and retrieval are also possible.

Consider the 2D example in Figure~\ref{fig:boundary_trees_intro}. Samples from the green shaded area are labelled 0 (colored green) and samples from the white area are labelled 1 (colored red). Boundary trees are constructed in an online manner, sample by sample. Each node of the boundary tree corresponds to a sample from training data. The tree is initialized (setting the root $\x_{root}$) with the first sample and its associated label (label 1, in this case, Figure~\ref{fig:boundary_trees_intro}). Given a new query sample $\y$ (in blue), we greedily traverse the tree: starting at the root, we find the closest node (according to some distance function) out of the children of the current node and the node itself, and recursively continue that traversal until we either reach a leaf (a node with no children) or stay at the current node (which means that it is closer to the query point than any of its immediate children). 
We use the label associated with the final node to produce the prediction of the tree (Figure~\ref{fig:boundary_trees_intro}); in this case, the prediction would be red. 
If the prediction is wrong (that is, the label associated with the final node is different than label of the query point) we add a new node containing the query point as a child to the final node (Figure~\ref{fig:boundary_trees_intro}). If the prediction is right, we discard the query node.

The resulting tree has the following interesting property -- every edge, by definition, crosses a boundary between the classes, and the nodes stored in the tree will tend to reside close to these boundaries (Figure~\ref{fig:boundary_trees_intro}). This means that a good representation should try to separate the
classes and form simple boundaries between them, resulting in efficient tree structures.

\subsection{Differentiable boundary trees}

Now that we have an understanding of how boundary trees work we ask the question: is there a way to learn a good representation for boundary trees? Such a representation should be one that transforms the samples in a way which transforms the boundaries between different classes to be as simple as possible. This would result in simpler tree structures, faster queries and less memory requirements.

One way to achieve this is to associate a differentiable cost function with a boundary tree and optimize it directly. In order to achieve this we propose to model transitions in the tree as stochastic events where the respective probabilities are a function of the distance between the query node and the nodes in the tree.

Let $\x$ denote the features of a data point, and let $c$ denote one-hot encoding of the associated class label.
Given the current node $\x_i$ and the query node $\y$ we model the transition probability to node $\x_j$, where $\x_j 
\in \{\child(\x_i),\x_i\}$, that is one of the children of node $\x_i$ or staying at this node, as the following: 
\BEA
p(\x_i\to \x_j|\y)=\underset{i,j\in \child(i)}{\mathsf{SoftMax}}(-d(\x_j,\y)) \nonumber\\
=\frac{\exp(-d(\x_j,\y))}{\sum_{j'\in\{{i,j\in \child(i)}\}}\exp(-d(\x_{j'},\y))}
\EEA
Where $d(\x, \y)$ is a distance function between $\x$ and $\y$. Though there are a wide variety of possible choices for $d$, we set it to be the $\Ltwo$ distance for the remainder of the paper:
\BE
d(\x,\y) = \sqrt{\sum_k (x_k - y_k)^2}
\EE
where we note that $\x$ can be an embedding (as we show below, after obtaining a differentiable cost function we can augment it with a neural network to produce these embeddings) and not necessarily raw data.

\begin{figure}[t]
\centering
\includegraphics[width=0.9\linewidth]{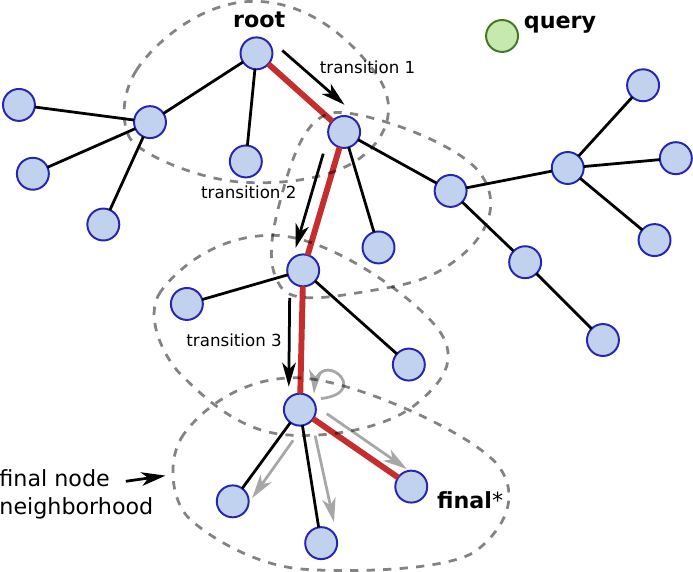}
\caption{\small Visualization of the different neighborhoods and transitions involved in the construction of the cost function in Equation \ref{eq:cost_function}. The tree is presented in an arbitrary 2D space here (for visualization). Given the query node we greedily traverse the tree down $\treepath^*$ (marked in red) after transforming all samples through $\ftheta$. The probability of each transition is calculated up until the final node's neighborhood. Here we aggregate the nodes' class labels, weighted by their respective transition probability, to build the class prediction output. See Figure \ref{fig:tree_nn} for a visualization of the associated neural net which is used to compute the cost function. }
\label{fig:tree_path}
\end{figure}

A path conditioned upon a query node $\y$, denoted $\treepath(\y)$, travels from the root node to the final node is a series of transitions $i\to j$ in the tree, each conditioned on the query node. Each transition is conditionally independent of the previous transitions. The probability of a path given a query node $\y$ is thus the product of probabilities for each transition along the path:
\BE
p(\treepath|\y)=\kern-1em\prod_{i \to j\in \treepath}\kern-1em p(\x_i\to \x_j|\y)
\EE
Finally, the predicted class $c$ probability distribution given the tree and a query node $p(c|\y)$ is the expected prediction of final nodes over all possible paths. In practice, calculating the full expected class distribution may be quite expensive so we approximate this by a greedily chosen path
($\treepath^*$) which is the same path chosen under the boundary tree algorithm:
\BE
p(c|\y)=\mathbb{E}_{\treepath|\y}(p(c|\treepath,\y))\approx p(c|\treepath^*,\y)
\EE
Instead of using just the final node as the prediction we can use all the nodes that participated in the final transition in building the output so we get softer class predictions. We remove the last transition from $\treepath^*$ to obtain $\treepath^\dagger$. The final class log probabilities
from the tree given the query node is:
\begin{align}
&\log p(c|\y) = \kern-2em \sum_{\x_i\to \x_j \in \treepath^\dagger|\y}\kern-2em \log p(\x_i\to \x_j|\y) \nonumber\\
&\qquad\qquad +  \log \kern-2em \sum_{\x_k \in \sieb(\x_{\leaf^*})}\kern-2em p(\parent(\x_k)\to \x_k|\y)c(\x_k)
\end{align}
where $\sieb(\x_i)$ are all the nodes sharing a parent with node $\x_i$ and the node $\x_i$ itself (because the algorithm may stop at a non-leaf node), $c(\x_i)$ is an indicator function for the class associated with node $\x_i$ (a ``one hot" encoding vector) and $\x_{\leaf^*}$ is the final node in the greedy path. We normalize the class probabilities at the output to obtain a proper distribution. See Figure \ref{fig:tree_path} for a visualization of the different elements.

Now, instead of using the samples in their raw representation, we can transform them using a deep neural net $f_{\theta}(x)$ (the “transform”)
such that we can learn a better representation of the data. This yields the following log class probabilities:
\BEA
\log p(c|\ftheta(\y)) = \kern-3em \sum_{\x_i\to \x_j \in \treepath^\dagger|\ftheta(\y)}\kern-2.5em 
\log p\left( \ftheta(\x_i) \to \ftheta(\x_j)|\ftheta(\y)\right) + \nonumber\\
\kern-3em    \log \kern-2em \sum_{\x_k \in \sieb(\x_{\leaf^*})}\kern-2em p(\parent(\ftheta(\x_k))\to \ftheta(\x_k)|\ftheta(\y))c(\x_k)
\label{eq:cost_function}
\EEA
Plugging Equation \ref{eq:cost_function} into a loss function (we use cross-entropy loss but other choices are possible) we can perform back-propagation in order to learn the parameters $\theta$ of the transform. 

All of these calculations assume that the tree structure remains fixed -- see Section \ref{sec:optimization} of how we handle this requirement in practice.

\subsection{Building the neural net}

\begin{figure*}[t]
\centering
\includegraphics[width=0.9\linewidth]{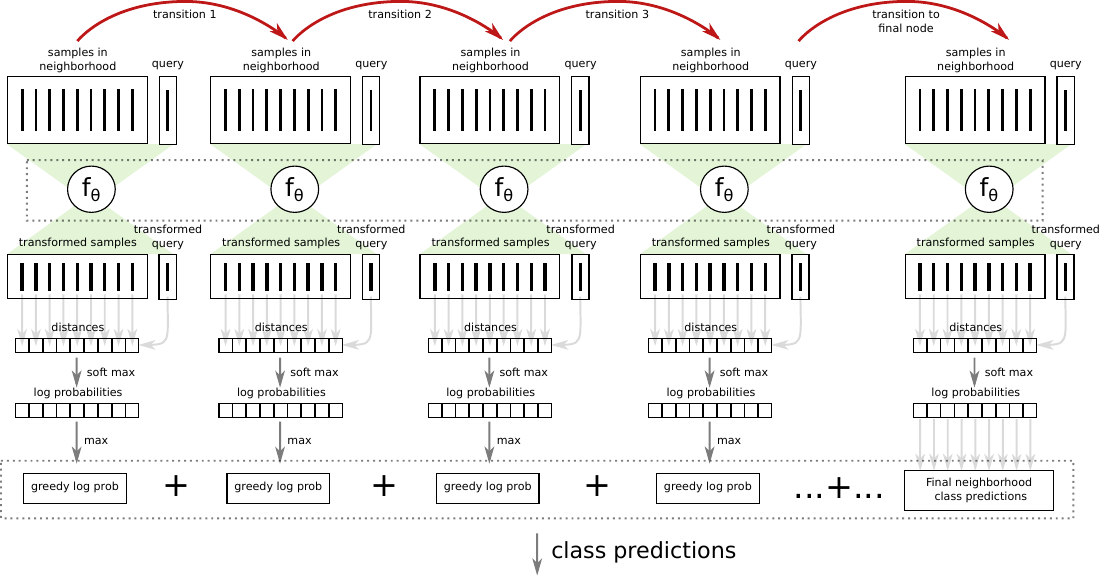}
\caption{\small The neural net needs to be dynamically constructed for each query point. For each transition through the tree modules outputting the transformed samples are shared. Each module takes the transformed samples, calculates distances between them and the transformed query point, then converts to log probabilities. Transitions are based on the transformed samples. These are combined with the final node's predictions to produce the class predictions, and the loss is propagated through the built network. See Figure  \ref{fig:tree_path} for a visualization of the corresponding tree structure and path.}
\label{fig:tree_nn}
\end{figure*}

For each query point, we need to dynamically build a neural network which corresponds to the chosen path in the tree, get its class predictions and loss and then calculate its gradient w.r.t the parameters of the transform $f_{\theta}$. Figure \ref{fig:tree_nn} depicts the structure of the network for an arbitrary path. 

One thing to note is that each query point results in a different network being built, but the parameters of the transform $f_{\theta}$ are shared throughout the process.

\subsection{Optimization}
\label{sec:optimization}
Since construction of the boundary tree requires discrete steps such as node and edge manipulation in the graph, it does not yield itself to back-propagation easily. In order to tackle this we choose to optimize the model in an iterative manner: we first build a boundary tree with a small number of samples, each transformed using the current transform $\ftheta$. Then, using the learned tree and keeping it fixed, we take several gradient steps w.r.t $\theta$ to update the representation. We repeat this process until some convergence criteria is met (in our case, when loss changes less than a specific threshold). We use Adam \citep{kingma2014adam} as the gradient descent optimizer , though other stochastic gradient descent approaches are also possible. See Algorithm~\ref{alg:pseudocode} for a summary.

Throughout all experiments we use a single tree, not a forest. It is possible to use a forest but we found that this did not make a considerable difference to learning and it is slower (data not shown).It is worth noting that the ordering of the samples presented to the tree can have a significant effect on its performance but over the course of training this effect is subdues as the representation is improved (See below).

\begin{algorithm*}[ht]
\caption{Learning representation using differentiable boundary trees}
\label{alg:pseudocode}
\begin{algorithmic}[1]
\State Initialize $\ftheta$ to random weights 
\While{not converged} \algcomment{Convergence when change is below a threshold}
\State Discard current tree $T$ and initialize new tree
\State Train new tree $T$ using samples transformed with current $\ftheta$
\For {$t=1, \mathsf{NumSamplesForRepresentations}$}
\State Get next training sample $\y$
\State Calculate loss using Equation \ref{eq:cost_function} and current tree $T$
\State Take gradient of loss w.r.t parameters $\theta$ and perform gradient step
\EndFor
\EndWhile
\end{algorithmic}
\end{algorithm*}

\section{Related work}
\label{sec:related_work}
There is a large body of literature on \knn\  methods and representation learning. Here we will mention some of the works which are most relevant to this one, but many others exist.

There are different families of nearest neighbors such as tree-based methods, hashing-based methods, etc. Our work falls under the former category. Tree-based methods can be further subdivided into methods that recursively partition input space using splits (e.g. $k$-d trees \citep{friedman1977algorithm} and random forests \citep{breiman2001random}) and methods that rely on distance comparisons to traverse down the tree. Examples of the latter include algorithms such as cover trees \citep{beygelzimer2006cover}, ball trees \citep{liu2006new}, boundary trees \citep{BF} and our proposed extension.

One of the key components in our method is modelling transition through the tree as stochastic events. Similar stochastic decisions have been explored in the decision literature; for example hierarchical mixture of experts \citep{HME} and more recently, the work on so-called \emph{deep neural decision forests} \citep{deepNDF}. Though it shares some of the basic ideas, the latter is fundamentally different from our method --- in the tree decision process (features vs. samples), in the tasks solved (directly solving classification vs. representation learning) and in the optimization process.

On the representation learning side, there are several works which are quite close to this one. In \citep{kochsiamese} a Siamese network is built to solve a `same' versus 'not same' labelling task. The network is built with two streams, each receiving a sample which undergoes under some transformation (where the weights are shared between the two streams). The task for the net is to say whether the two samples are from the same class. On some level, this can be seen as a special case of our method where the tree consists of exactly two nodes and the loss function is an indicator function for class similarity. Another related representation learning method which is relevant is \citep{hoffer2015deep}. In this work a network is trained to predict class similarity using three samples: one sample is the reference sample, one comes from the same class as the reference and the other comes from a different class. The goal of the network is to push samples from the same class closer together and samples from different classes apart. Again, the weights are shared between the different streams of the system. This is can also seen as another special case of our method where the structure of the tree and samples chosen in its construction are pre-set and with a somewhat different loss function.

Two particularly related works are \citep{goldberger2004neighbourhood} and \citep{craven1996extracting}. In \citep{goldberger2004neighbourhood} a Mahalanobis distance metric is learned with the objective of improving nearest neighbor classification. Our work can be seen as a deep non-linear version of this work, augmented by a more efficient and structured nearest neighbor method. Finally, in \citep{craven1996extracting} a decision tree is built using trained neural net features in order to obtain an interpretable decision structure. Our work is related to this idea, though we provide an end-to-end solution which learns the neural net \emph{together} with the tree structure.

\section{Experiments and analysis}

\subsection{Half-moon dataset}

\begin{figure}[t]
\centering
\includegraphics[width=0.5\linewidth]{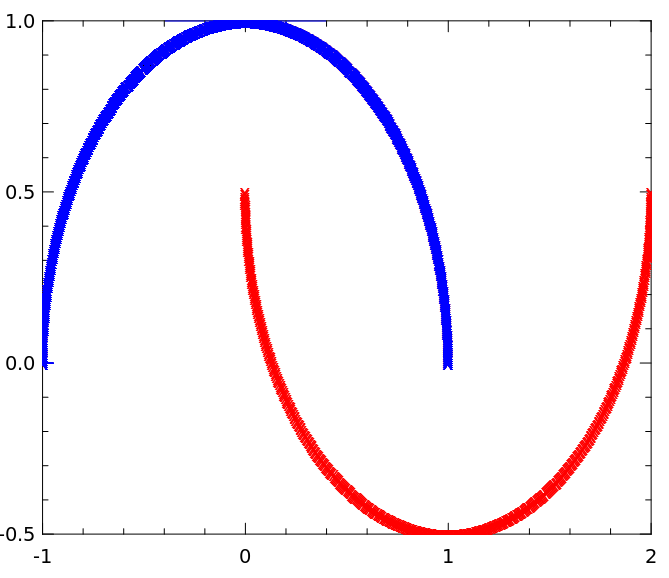}%
\includegraphics[width=0.5\linewidth]{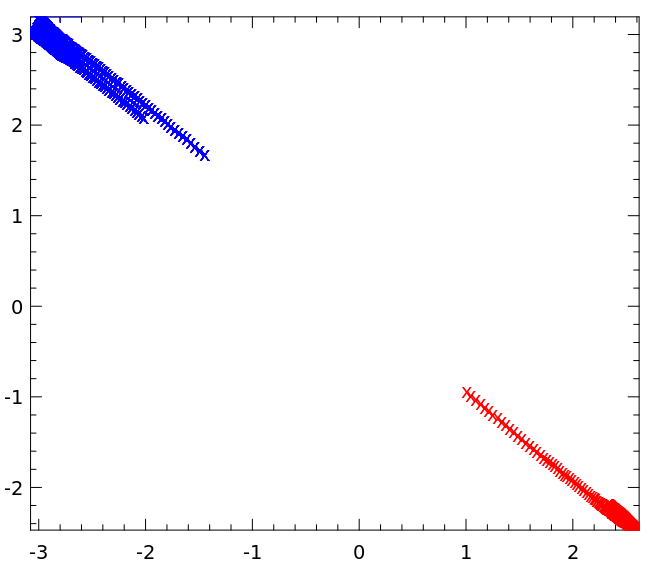}\\
\caption{\small Half moon dataset and the learned transformation. On the left is the raw data, on the right is the transformed data using a learned representation. We train a simple 3 hidden layer fully connected net with our method. As can be seen, the network learns to separate the two classes completely. Using this representation, a trained boundary tree has only 2 nodes and achieves 0\% error on both train and test sets.}
\label{fig:half_moon}
\vspace{-1em}
\end{figure}

To gain an intuition about the workings of the our method we start with the \emph{half-moons} dataset. This dataset has two classes, each lying on a one dimensional manifold on a 2D plane (Figure~\ref{fig:half_moon}). With enough samples, this is a task which is easy to solve with a variety of methods -- including nearest neighbors based methods such as the boundary forest. Looking at Figure~\ref{fig:half_moon}, it is clear that a non-linear transformation of the data may allow for a much simpler solution.

We learn a representation using our method --- training on 1000 samples, iteratively building a boundary tree using 20 samples, and taking 10 gradient steps (each step over a different new sample not used in the tree) to update the representation $f_{\theta}$ using the learned tree, repeating until convergence. For the transform $\ftheta$ we train a 3 layer, fully connected net ($2\to 100\to100\to 30\to 2$ units). Figure~\ref{fig:half_moon} shows the transformed samples after we train the representation using our method. Indeed, the transformation has learned to separate the two classes into two distinct clusters. When using this representation, training a boundary tree results in a 2 node tree which yields 0\% training and 0\% test errors. 

\subsection{MNIST classification}

To test our method on a more ``real world" dataset (albeit a simple one) we use the MNIST handwritten digit dataset. At each iteration we build a tree using 1000 samples transformed with current transform $\ftheta$,  then take a 1000 gradient steps (each on one new sample) to update the representation. We repeat this process, rebuilding the tree at each iteration until convergence. The representation is a 2 hidden layer, fully connected net ($784\to 400\to 400\to 20$ units respectively). We use Adam \citep{kingma2014adam} as our back-propagation optimizer. No pre-processing or data augmentation is used. 

\begin{table*}[t]

  \centering
  \begin{tabular}{lll}
    \toprule
    \textbf{Method} & \textbf{Test Error Rate} & \textbf{Number of nodes}\\ 
    \midrule
	Boundary tree (raw pixels) & 11.01\% & 8536\\
	Boundary tree (pre-trained net) & 5.5\% & 2906\\
	$1$-NN (Raw pixels) & 5.0\% & 60,000\\
	Neural net (directly as classfier) & 2.4\% & -\\
	\midrule
    Boundary tree (our learned representation) & \textbf{1.85\%} & \textbf{202}\\
    \bottomrule
  \end{tabular}
    \caption{Results on MNIST classification. The trees were built over the entire 60,000 sample training set. All results are reported for a single tree (not with forests). Note that with our learned representation only a small handful of samples is needed. The neural net used is a fully connected net as described in the text.} 
  \label{tab:MNIST}
\end{table*}

\begin{figure}[htbp]
\centering
\includegraphics[width=0.985\linewidth]{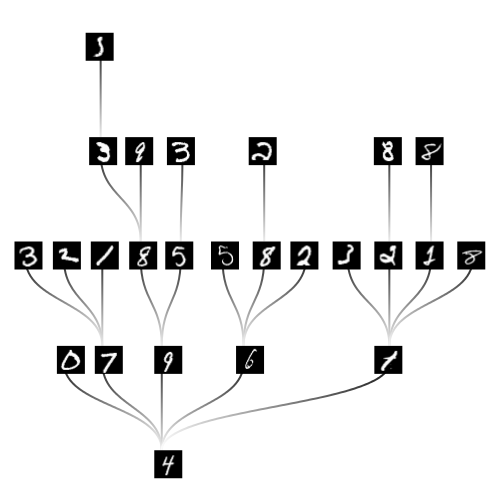}
\caption{\small A tree built using the trained representation over MNIST digits using 1000 samples. Samples are presented here in the original pixel space, but the learned representation is used to construct the tree. Note the simple, interpretable structure --- nodes are either prototypical examples or boundary cases. Remarkably, this tree still achieves less than 2\% error on the test set. See text for details.}
\label{fig:mnist_tree}
\end{figure}

\begin{figure*}[t]
\centering
\includegraphics[width=0.455\linewidth]{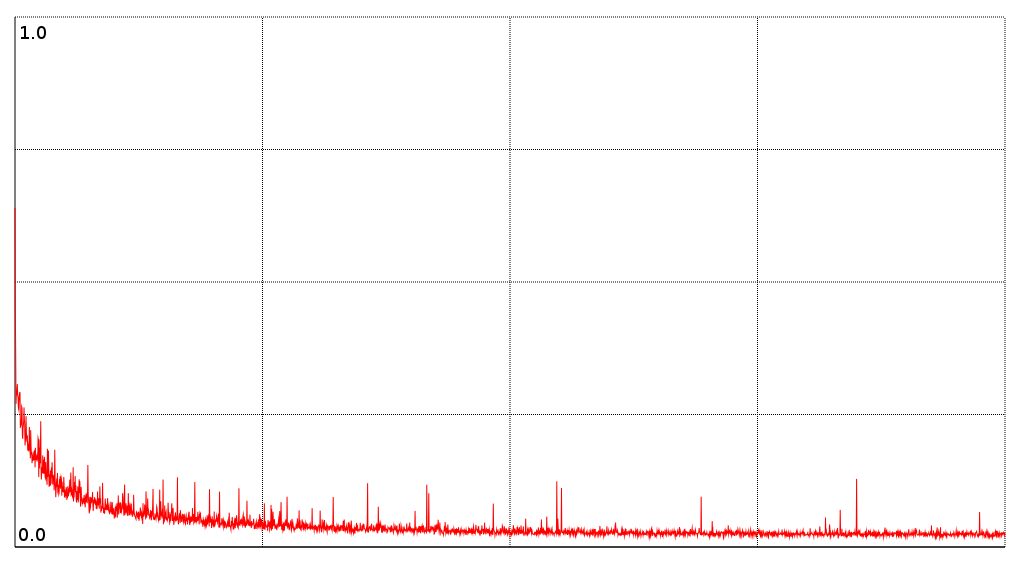}\quad\quad
\includegraphics[width=0.455\linewidth]{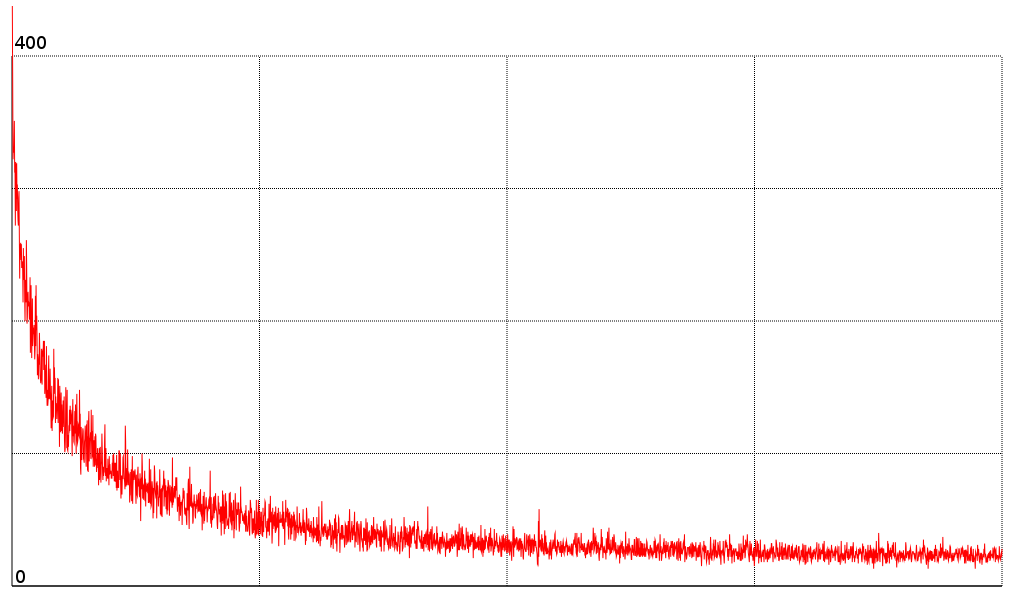}%
\caption{%
Test error (left) and number of nodes in the tree (right) as a function of iterations. The tree was built using 1000 samples, and then, keeping the tree fixed the representation was updated over another 1000 gradient steps, repeating this. Note that as learning progresses the tree needs to keep less and less nodes. By the end of learning only about 25 nodes are needed to achieve less than 2\% test error.}
\label{fig:mnist_learning}
\end{figure*}

At the end of training we obtain an extremely simple tree with which we make the classification decisions. This simple tree is displayed in Figure~\ref{fig:mnist_tree}. Samples are shown in their original pixel space (though the transformed representation is used to make the traversal decisions). As can be seen, the tree has a very interpretable structure --- samples are either prototypical (for some of the classes such as ``4" a single example is all that is needed) or more esoteric boundary cases (cropped or distorted digits). This is one of the more appealing properties of our method --- we can actually try and understand what is the decision process the tree takes, in contrast to other tree based methods \citep{breiman2001random, deepNDF}. Remarkably, this tree still achieves less than 2\% error on the test set. 

Full results on the test set are presented in Table \ref{tab:MNIST}. As can be seen, using the representation yields better results than using raw-pixels (as in \citep{BF}) with a single tree. We also find that training a network with a comparable architecture directly on the classification task yields a representation which is less suited for a \knn\  method such as the boundary tree (second row in the table).

Figure~\ref{fig:mnist_learning} shows the test error and number of nodes in the tree as a function of iterations. As can be clearly seen, as the representation improves, the tree needs to keep less and less nodes --- by the end of learning, the tree has just about 25 nodes, still achieving below 2\% error on the test set.

\begin{figure*}[b]
\centering
\includegraphics[width=0.455\linewidth]{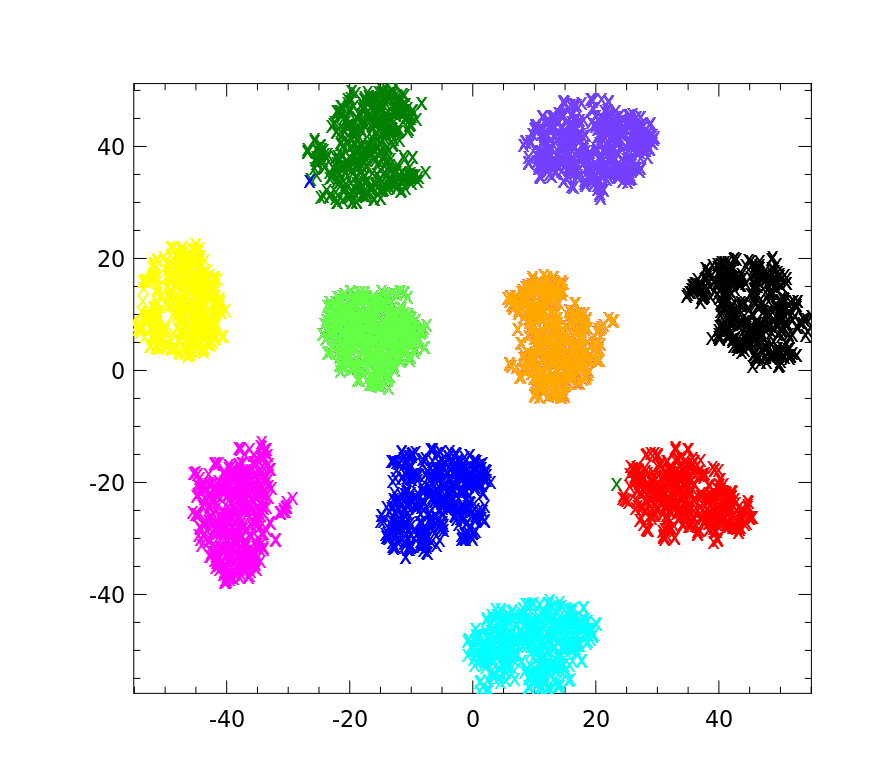}%
\includegraphics[width=0.455\linewidth]{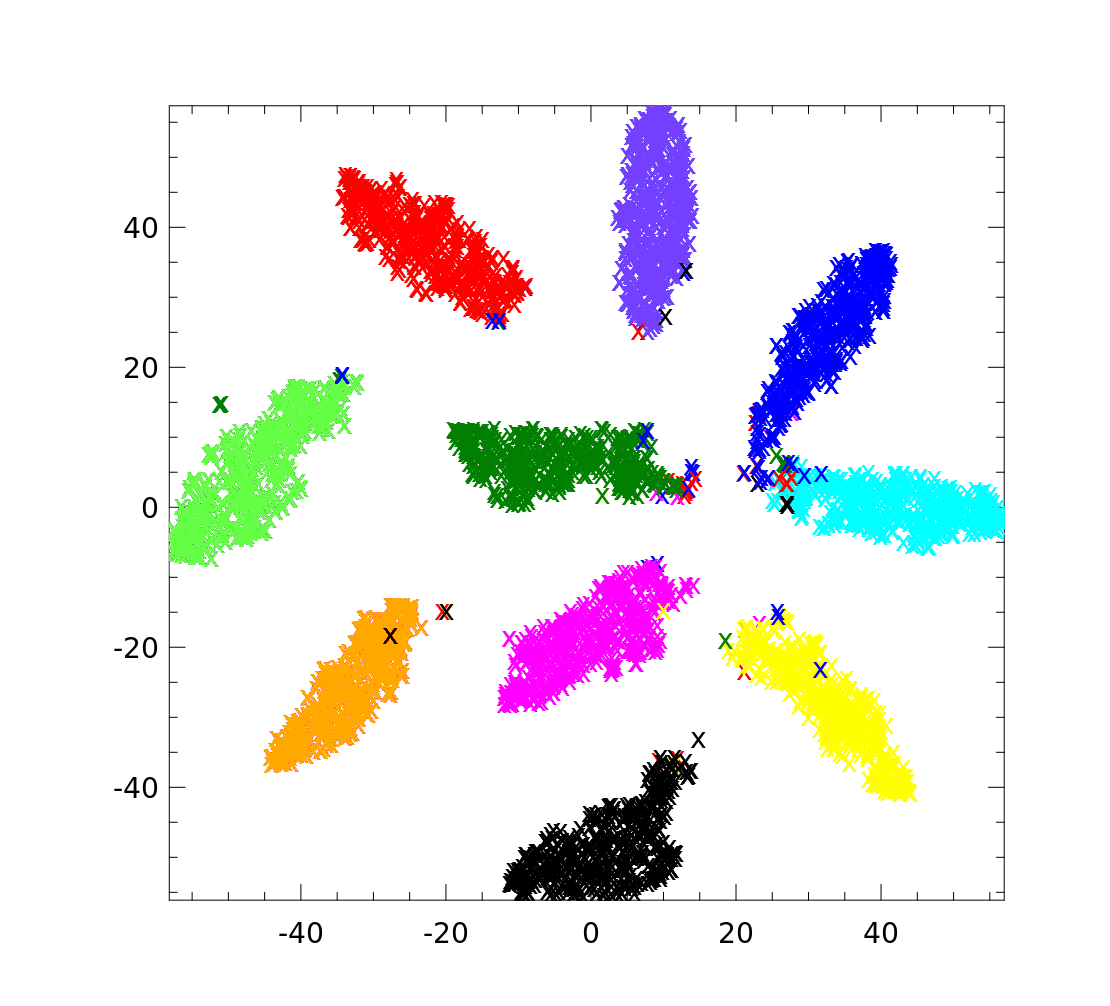}%
\caption{%
$t$-SNE visualisation of representations. On the left are samples projected using our trained representation (from the training set). As can be seen, the transform separates the classes nicely, allowing the tree to have a very small number of nodes in order to achieve its task. On the right is the t-SNE plot made with neural net classifier outputs. Though classes are separated nicely, it's not as clean as our result. 
}
\label{fig:mnist_tsne}
\end{figure*}

How can the tree store such a small number of samples and still achieve this level of performance? In order to understand what the representation is doing, we plotted a $t$-SNE \citep{van2008visualizing} visualisation for the learned 20 dimensional representation, projected down to 2D. Figure~\ref{fig:mnist_tsne} shows the results, together with the result of a pre-trained MNIST network of similar structure (trained directly on classification). As can be seen, the representation we learn clearly separates the classes from one another, much more than the neural net directly trained to classify. This enables the simple tree structure which is learned after transformation.

\subsection{Limitations and scaling}

Due to the discrete nature of the path we take through the tree we need to build a different computation graph for each query node. This makes batching very inefficient and thus prevents us from running experiments on larger scale data such as ImageNet images requiring the use of GPUs. Another limitation is the iterative nature of the algorithm which requires switching between building the tree (a discrete operation) and updating the representation (which is continuous) --- it would be more elegant and more efficient to perform both under the same framework.

Another limitation is that early in training the resulting tree may be quite large as the tree frequently errs, many samples are added. This may be alleviated by initially using less samples for building the tree increasing the variance of gradients but reducing computational and memory requirements.

\begin{figure}[H]
\centering
\includegraphics[width=0.985\linewidth]{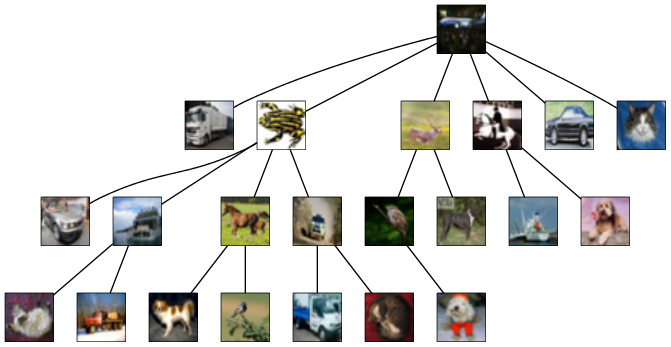}
\caption{\small A tree built using the pre-trained representation over CIFAR10 images using 100 samples. Samples are presented here in the original pixel space, but the learned representation is used to construct the tree. Note the simple, interpretable structure --- nodes are either prototypical examples or boundary cases. This tree achieves 13.06\% error on the test set using just 22 samples. See text for details.}
\label{fig:cifar_tree}
\end{figure}

Though we cannot learn the representation directly on larger datasets we can certainly try to visualize what a boundary tree trained on pre-trained representation would yield. Figure \ref{fig:cifar_tree} shows an example of this: we used a pre-trained VGG-style network which was optimized for predicting classes on the CIFAR10 dataset using the cross entropy loss. We then 
use the outputs from the layer before the softmax, as embeddings for the inputs.  
As can be seen, the tree learned from 100 samples is quite compact with 22 nodes. The classification error for the tree is 13.06\% vs. 9\% for the pre-trained net.

\section{Discussion}
We have presented a method to learn useful representations for \knn\  methods. Using boundary trees we are able to derive a differentiable cost function which allows for learning of such representations. The resulting representation allows for simple, interpretable tree structures with good performance in query time and accuracy. While the method has some limitations we feel this is an important research direction and are looking forward to further explore this directions using newly available dynamic batching tools such as TensorFlow Fold.\footnote{\url{https://github.com/tensorflow/fold}}

\clearpage
\bibliography{paper}
\bibliographystyle{abbrvnat}

\end{document}

%% file: commands.tex
\newcommand{\y}{\mathbf{y}}

\newcommand{\Ltwo}{\mathbb{L}_2}
\renewcommand{\vec}[1]{\mathbf{#1}}

\newcommand{\x}{\vec{x}}

\newcommand{\future}[1]{} %
\newcommand{\comment}[1]{}

\newcommand{\later}[1]{}
\newcommand{\longer}[1]{}	%

\def\BE{\begin{equation}}
\def\EE{\end{equation}}

\def\BEA{\begin{eqnarray}}
\def\EEA{\end{eqnarray}}
\def\BEAS{\begin{eqnarray*}}
\def\EEAS{\end{eqnarray*}}